# Machine learning and deep learning

Christian Janiesch[1] · Patrick Zschech[2] · Kai Heinrich[3]

## Abstract
Today, intelligent systems that offer artificial intelligence capabilities often rely on machine learning. Machine learning describes the capacity of systems to learn from problem-specific training data to automate the process of analytical model building and solve associated tasks. Deep learning is a machine learning concept based on artificial neural networks. For many applications, deep learning models outperform shallow machine learning models and traditional data analysis approaches. In this article, we summarize the fundamentals of machine learning and deep learning to generate a broader understanding of the methodical underpinning of current intelligent systems. In particular, we provide a conceptual distinction between relevant terms and concepts, explain the process of automated analytical model building through machine learning and deep learning, and discuss the challenges that arise when implementing such intelligent systems in the field of electronic markets and networked business. These naturally go beyond technological aspects and highlight issues in human-machine interaction and artificial intelligence servitization.

**Keywords** Machine learning · Deep learning · Artificial intelligence · Artificial neural networks · Analytical model building

**JEL classification** C6 · C8 · M15 · O3

## Introduction

It is considered easier to explain to a child the nature of what constitutes a sports car as opposed to a normal car by showing him or her examples, rather than trying to formulate explicit rules that define a sports car.

Similarly, instead of codifying knowledge into computers, machine learning (ML) seeks to automatically learn meaningful relationships and patterns from examples and observations (Bishop 2006). Advances in ML have enabled the recent rise of intelligent systems with human-like cognitive capacity that penetrate our business and personal life and shape the networked interactions on electronic markets in every conceivable way, with companies augmenting decision-making for productivity, engagement, and employee retention (Shrestha et al. 2021), trainable assistant systems adapting to individual user preferences (Fischer et al. 2020), and trading agents shaking traditional finance trading markets (Jayanth Balaji et al. 2018).

The capacity of such systems for advanced problem solving, generally termed artificial intelligence (AI), is based on analytical models that generate predictions, rules, answers, recommendations, or similar outcomes. First attempts to build analytical models relied on explicitly programming known relationships, procedures, and decision logic into intelligent systems through handcrafted rules (e.g., expert systems for medical diagnoses) (Russell and Norvig 2021). Fueled by the practicability of new programming frameworks, data availability, and the broad access to necessary computing power, analytical models are nowadays increasingly built using what is generally referred to as ML (Brynjolfsson and McAfee 2017; Goodfellow et al. 2016). ML relieves the human of the burden to explicate and formalize his or her

Springer

knowledge into a machine-accessible form and allows to develop intelligent systems more efficiently.

During the last decades, the field of ML has brought forth a variety of remarkable advancements in sophisticated learning algorithms and efficient pre-processing techniques. One of these advancements was the evolution of artificial neural networks (ANNs) towards increasingly deep neural network architectures with improved learning capabilities summarized as deep learning (DL) (Goodfellow et al. 2016; LeCun et al. 2015). For specific applications in closed environments, DL already shows superhuman performance by excelling human capabilities (Madani et al. 2018; Silver et al. 2018). However, such benefits also come at a price as there are several challenges to overcome for successfully implementing analytical models in real business settings. These include the suitable choice from manifold implementation options, bias and drift in data, the mitigation of black-box properties, and the reuse of preconfigured models (as a service).

Beyond its hyped appearance, scholars, as well as professionals, require a solid understanding of the underlying concepts, processes as well as challenges for implementing such technology. Against this background, the goal of this article is to convey a fundamental understanding of ML and DL in the context of electronic markets. In this way, the community can benefit from these technological achievements – be it for the purpose of examining large and high-dimensional data assets collected in digital ecosystems or for the sake of designing novel intelligent systems for electronic markets. Following recent advances in the field, this article focuses on analytical model building and challenges of implementing intelligent systems based on ML and DL. As we examine the field from a technical perspective, we do not elaborate on the related issues of AI technology adoption, policy, and impact on organizational culture (for further implications cf. e.g. Stone et al. 2016).

In the next section, we provide a conceptual distinction between relevant terms and concepts. Subsequently, we shed light on the process of automated analytical model building by highlighting the particularities of ML and DL. Then, we proceed to discuss several induced challenges when implementing intelligent systems within organizations or electronic markets. In doing so, we highlight environmental factors of implementation and application rather than viewing the engineered system itself as the only unit of observation. We summarize the article with a brief conclusion.

## Conceptual distinction

To provide a fundamental understanding of the field, it is necessary to distinguish several relevant terms and concepts from each other. For this purpose, we first present basic foundations of AI, before we distinguish i) machine learning algorithms, ii) artificial neural networks, and iii) deep neural networks. The hierarchical relationship between those terms is summarized in Venn diagram of Fig. 1.

Broadly defined, AI comprises any technique that enables computers to mimic human behavior and reproduce or excel over human decision-making to solve complex tasks independently or with minimal human intervention (Russell and Norvig 2021). As such, it is concerned with a variety of central problems, including knowledge representation, reasoning, learning, planning, perception, and communication, and refers to a variety of tools and methods (e.g., case-based reasoning, rule-based systems, genetic algorithms, fuzzy models, multi-agent systems) (Chen et al. 2008). Early AI research focused primarily on hard-coded statements in formal languages, which a computer can then automatically reason about based on logical inference rules. This is also known as the knowledge base approach (Goodfellow et al. 2016). However, the paradigm faces several limitations as humans generally struggle to explicate all their tacit knowledge that is required to perform complex tasks (Brynjolfsson and McAfee 2017).

*Machine learning* overcomes such limitations. Generally speaking, ML means that a computer program's performance improves with experience with respect to some class of tasks and performance measures (Jordan and Mitchell 2015). As such, it aims at automating the task of analytical model building to perform cognitive tasks like object detection or natural language translation. This is achieved by applying algorithms that iteratively learn from problem-specific training data, which allows computers to find hidden insights and complex patterns without explicitly being programmed (Bishop 2006). Especially in tasks related to high-dimensional data such as classification, regression, and clustering, ML shows good applicability. By learning from previous computations and extracting regularities from massive databases, it can help to produce reliable and repeatable decisions. For this reason, ML algorithms have been successfully applied in many areas, such as fraud detection, credit scoring, next-best offer analysis, speech and image recognition, or natural language processing (NLP).

Based on the given problem and the available data, we can distinguish three types of ML: supervised learning, unsupervised learning, and reinforcement learning. While many applications in electronic markets use supervised learning (Brynjolfsson and McAfee 2017), for example, to forecast stock markets (Jayanth Balaji et al. 2018), to understand customer perceptions (Ramaswamy and DeClerck 2018), to analyze customer needs (Kühl et al. 2020), or to search products (Bastan et al. 2020), there are implementations of all types, for example, market-making with reinforcement learning (Spooner et al. 2018) or unsupervised market segmentation using customer reviews (Ahani et al. 2019). See Table 1 for an overview of all three types.

**Fig. 1** Venn diagram of machine learning concepts and classes (inspired by Goodfellow et al. 2016, p. 9)

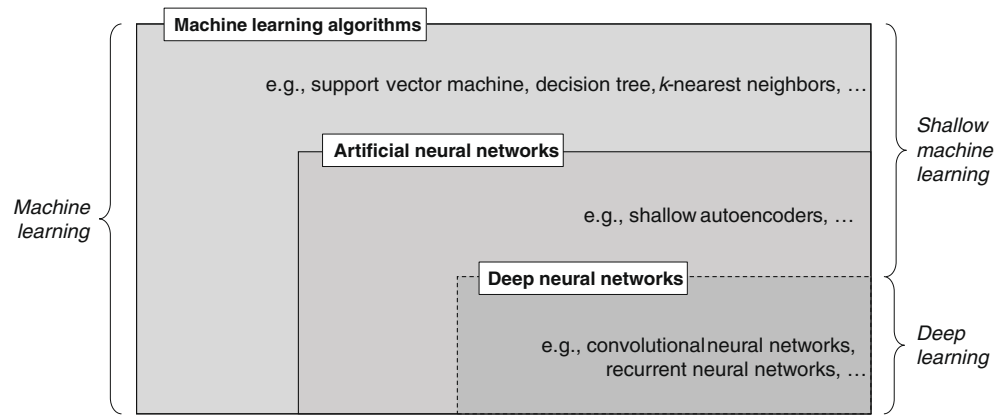

Depending on the learning task, the field offers various classes of *ML algorithms*, each of them coming in multiple specifications and variants, including regressions models, instance-based algorithms, decision trees, Bayesian methods, and ANNs.

The family of *artificial neural networks* is of particular interest since their flexible structure allows them to be modified for a wide variety of contexts across all three types of ML. Inspired by the principle of information processing in biological systems, ANNs consist of mathematical representations of connected processing units called artificial neurons. Like synapses in a brain, each connection between neurons transmits signals whose strength can be amplified or attenuated by a weight that is continuously adjusted during the learning process. Signals are only processed by subsequent neurons if a certain threshold is exceeded as determined by an activation function. Typically, neurons are organized into networks with different layers. An input layer usually receives the data input (e.g., product images of an online shop), and an output layer produces the ultimate result (e.g., categorization of products). In between, there are zero or more hidden layers that are responsible for learning a non-linear mapping between input and output (Bishop 2006; Goodfellow et al. 2016). The number of layers and neurons, among other property choices, such as learning rate or activation function, cannot be learned by the learning algorithm. They constitute a model's hyperparameters and must be set manually or determined by an optimization routine.

*Deep neural networks* typically consist of more than one hidden layer, organized in deeply nested network architectures. Furthermore, they usually contain advanced neurons in contrast to simple ANNs. That is, they may use advanced operations (e.g., convolutions) or multiple activations in one neuron rather than using a simple activation function. These characteristics allow deep neural networks to be fed with raw input data and automatically discover a representation that is needed for the corresponding learning task. This is the networks' core capability, which is commonly known as *deep learning*. Simple ANNs (e.g., shallow autoencoders) and other ML algorithms (e.g., decision trees) can be subsumed under the term *shallow machine learning* since they do not provide such functionalities. As there is still no exact demarcation between the two concepts in literature (see also Schmidhuber 2015), we use a dashed line in Fig. 1. While some shallow ML algorithms are considered inherently interpretable by humans and, thus, white boxes, the decision making of most advanced ML algorithms is per se untraceable unless explained otherwise and, thus, constitutes a black box.

**Table 1** Overview of types of machine learning

| Type | Description |
| --- | --- |
| Supervised learning | *Supervised learning* requires a training dataset that covers examples for the input as well as labeled answers or target values for the output. An example could be the prediction of active users subscribed to a market platform in a month's time as output (considered as the target variable or *y* variable) based on different input characteristics, such as the number of sold products or positive user reviews (often referred to as input features or *x* variables). The pairs of input and output data in the training set are then used to calibrate the open parameters of the ML model. Once the model has been successfully trained, it can be used to predict the target variable *y* given new or unseen data points of the input features *x*. Regarding the type of supervised learning, we can further distinguish between regression problems, where a numeric value is predicted (e.g., number of users), and classification problems, where the prediction result is a categorical class affiliation such as "lookers" or "buyers". |
| Unsupervised learning | *Unsupervised learning* takes place when the learning system is supposed to detect patterns without any pre-existing labels or specifications. Thus, training data only consists of variables *x* with the goal of finding structural information of interest, such as groups of elements that share common properties (known as clustering) or data representations that are projected from a high-dimensional space into a lower one (known as dimensionality reduction) (Bishop 2006). A prominent example of unsupervised learning in electronic markets is applying clustering techniques to group customers or markets into segments for the purpose of a more target-group specific communication. |
| Reinforcement learning | In a *reinforcement learning* system, instead of providing input and output pairs, we describe the current state of the system, specify a goal, provide a list of allowable actions and their environmental constraints for their outcomes, and let the ML model experience the process of achieving the goal by itself using the principle of trial and error to maximize a reward. Reinforcement learning models have been applied with great success in closed world environments such as games (Silver et al. 2018), but they are also relevant for multi-agent systems such as electronic markets (Peters et al. 2013). |

DL is particularly useful in domains with large and high-dimensional data, which is why deep neural networks outperform shallow ML algorithms for most applications in which text, image, video, speech, and audio data needs to be processed (LeCun et al. 2015). However, for low-dimensional data input, especially in cases of limited training data availability, shallow ML can still produce superior results (Zhang and Ling 2018), which even tend to be better interpretable than those generated by deep neural networks (Rudin 2019). Further, while DL performance can be superhuman, problems that require *strong AI* capabilities such as literal understanding and intentionality still cannot be solved as pointedly outlined in Searle (1980)'s Chinese room argument.

## Process of analytical model building

In this section, we provide a framework on the process of analytical model building for explicit programming, shallow ML, and DL as they constitute three distinct concepts to build an analytical model. Due to their importance for electronic markets, we focus the subsequent discussion on the related aspects of data input, feature extraction, model building, and model assessment of shallow ML and DL (cf. Figure 2). With explicit programming, feature extraction and model building are performed manually by a human when handcrafting rules to specify the analytical model.

## Data input

Electronic markets have different stakeholder touchpoints, such as websites, apps, and social media platforms. Apart from common numerical data, they generate a vast amount of versatile data, in particular unstructured and non-cross-sectional data such as time series, image, and text. This data can be exploited for analytical model building towards better decision support or business automation purposes. However, extracting patterns and relationships by hand would exceed the cognitive capacity of human operators, which is why algorithmic support is indispensable when dealing with large and high-dimensional data.

Time series data implies a sequential dependency and patterns over time that need to be detected to form forecasts, often resulting in regression problems or trend classification tasks. Typical examples involve forecasting financial markets or predicting process behavior (Heinrich et al. 2021). Image data is often encountered in the context of object recognition or object counting with fields of application ranging from crop detection for yield prediction to autonomous driving (Grigorescu et al. 2020). Text data is present when analyzing large volumes of documents such as corporate e-mails or social media posts. Example applications are sentiment analysis or machine-based translation and summarization of documents (Young et al. 2018).

Recent advancements in DL allow for processing data of different types in combination, often referred to as cross-modal learning. This is useful in applications where content is subject to multiple forms of representation, such as e-commerce websites where product information is commonly represented by images, brief descriptions, and other complementary text metadata. Once such cross-modal representations are learned, they can be used, for example, to improve retrieval and recommendation tasks or to detect misinformation and fraud (Bastan et al. 2020).

## Feature extraction

An important step for the automated identification of patterns and relationships from large data assets is the extraction of features that can be exploited for model building. In general, a feature describes a property derived from the raw data input with the purpose of providing a suitable representation. Thus, feature extraction aims to preserve discriminatory information and separate factors of variation relevant to the overall learning task (Goodfellow et al. 2016). For example, when classifying the helpfulness of customer reviews of an online-shop, useful feature candidates could be the choice of words, the length of the review, and the syntactical properties of the text.

Shallow ML heavily relies on such well-defined features, and therefore its performance is dependent on a successful extraction process. Multiple feature extraction techniques have emerged over time that are applicable to different types of data. For example, when analyzing time-series data, it is common to apply techniques to extract time-domain features (e.g., mean, range, skewness) and frequency-domain features (e.g., frequency bands) (Goyal and Pabla 2015); for image analysis, suitable approaches include histograms of oriented gradients (HOG) (Dalal and Triggs 2005), scale-invariant feature transform (SIFT) (Lowe 2004), and the Viola-Jones method (Viola and Jones 2001); and in NLP, it is common to use term frequency-inverse document frequency (TF-IDF) vectors (Salton and Buckley 1988), part-of-speech (POS) tagging, and word shape features (Wu et al. 2018). Manual feature design is a tedious task as it usually requires a lot of domain expertise within an application-specific engineering process. For this reason, it is considered time-consuming, labor-intensive, and inflexible.

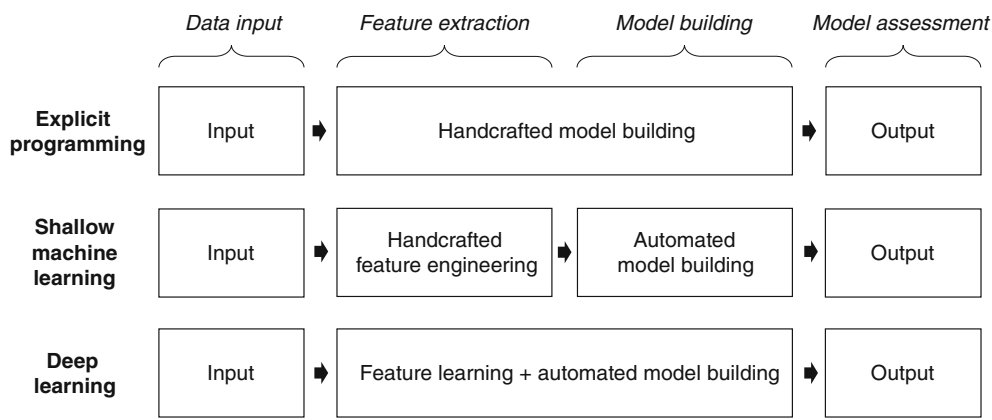

Fig. 2 Process of analytical model building (inspired by Goodfellow et al. 2016, p. 10)

Deep neural networks overcome this limitation of handcrafted feature engineering. Their advanced architecture gives them the capability of automated feature learning to extract discriminative feature representations with minimal human effort. For this reason, DL better copes with large-scale, noisy, and unstructured data. The process of feature learning generally proceeds in a hierarchical manner, with high-level abstract features being assembled by simpler ones. Nevertheless, depending on the type of data and the choice of DL architecture, there are different mechanisms of feature learning in conjunction with the step of model building.

## Model building

During automated model building, the input is used by a learning algorithm to identify patterns and relationships that are relevant for the respective learning task. As described above, shallow ML requires well-designed features for this task. On this basis, each family of learning algorithms applies different mechanisms for analytical model building. For example, when building a classification model, decision tree algorithms exploit the features space by incrementally splitting data records into increasingly homogenous partitions following a hierarchical, tree-like structure. A support vector machine (SVM) seeks to construct a discriminatory hyperplane between data points of different classes where the input data is often projected into a higher-dimensional feature space for better separability. These examples demonstrate that there are different ways of analytical model building, each of them with individual advantages and disadvantages depending on the input data and the derived features (Kotsiantis et al. 2006).

By contrast, DL can directly operate on high-dimensional raw input data to perform the task of model building with its capability of automated feature learning. Therefore, DL architectures are often organized as end-to-end systems combining both aspects in one pipeline. However, DL can also be applied only for extracting a feature representation, which is subsequently fed into other learning subsystems to exploit the strengths of competing ML algorithms, such as decision trees or SVMs.

Various DL architectures have emerged over time (Leijnen and van Veen 2020; Pouyanfar et al. 2019; Young et al. 2018). Although basically every architecture can be used for every task, some architectures are more suited for specific data such as time series or images. Architectural variants are mostly characterized by the types of layers, neural units, and connections they use. Table 2 summarizes the five groups of convolutional neural networks (CNNs), recurrent neural networks (RNNs), distributed representations, autoencoders, and generative adversarial neural networks (GANs). They provide promising applications in the field of electronic markets.

## Model assessment

For the assessment of a model's quality, multiple aspects have to be taken into account, such as performance, computational resources, and interpretability. Performance-based metrics evaluate how well a model satisfies the objective specified by the learning task. In the area of supervised learning, there are well-established guidelines for this purpose. Here, it is common practice to use $k$-fold cross-validation to prevent a model from overfitting and determine its performance on out-of-sample data that was not included in the training samples. Cross-validation provides the opportunity to compare the reliability of ML models by providing multiple out-of-sample data instances that enable comparative statistical testing (García and Herrera 2008). Regression models are evaluated by measuring estimation errors such as the root mean square error (RMSE) or the mean absolute percentage error (MAPE), whereas classification models are assessed by calculating different ratios of correctly and incorrectly predicted instances, such as accuracy, recall, precision, and $F1$ score. Furthermore, it is common to apply cost-sensitive measures such as average cost per predicted observation, which is helpful in situations where prediction errors are associated with asymmetric cost structures (Shmueli and Koppius 2011). That is the case, for example, when analyzing transactions in financial markets, and the costs of failing to detect a fraudulent transaction are remarkably higher than the costs of incorrectly classifying a non-fraudulent transaction.

**Table 2** Overview of deep learning architectures

| Architecture | Description |
| --- | --- |
| Convolutional neural network (CNN) | CNNs are mainly applied for tasks related to computer vision and speech recognition. They are able to address tasks involving datasets with spatial relationships, where the columns and rows are not interchangeable (e.g., image data). Their network architecture comprises a series of stages that allow hierarchical feature learning as determined by the respective modeling task. For example, when considering object recognition in images, the first few layers of the network are responsible for extracting basic features in the form of edges and corners. These are then incrementally aggregated into more complex features in the last few layers resembling the actual objects of interest, such as animals, houses, or cars. Subsequently, the auto-generated features are used for prediction purposes to recognize objects of interest in new images (Goodfellow et al. 2016). |
| Recurrent neural network (RNN) | RNNs are designed explicitly for sequential data structures such as time-series data, event sequences, and natural language. Their architecture offers internal feedback loops and therefore enables sequential pattern learning to model time dependencies by forming a memory. Simple RNN architectures are problematic since they suffer from vanishing gradients, resulting in little or no influence of early memories. More sophisticated architectures, such as long short-term memory (LSTM) networks with advanced attention mechanisms, attend to this problem. RNNs are typically applied for time series forecasting, predicting process behavior (Heinrich et al. 2021), and NLP tasks such as sequence transduction and neural machine translation (LeCun et al. 2015). |
| Distributed representation | Distributed representations play an essential role in feature learning and language modeling in NLP tasks, where language entities such as words, phrases, and sentences are projected into numerical representations within a unified semantic space in the form of embeddings. Word embeddings, for example, encode discrete words into dense feature vectors with low dimensionality. Thus, in contrast to classic text representation models, such as one-hot encodings and bag-of-words (BoW), word embeddings overcome the problem of sparse encodings while preserving semantic relationships between words. This means that words, which occur in similar contexts in a corpus, are also closely positioned to each other in the vector space. On this basis, advanced language models can be developed to perform challenging downstream tasks, such as question-answering, sentiment analysis, and named entity recognition (Liu et al. 2020). Distributed representations are often applied in combination with RNNs to perform tasks with sequential dependencies. |
| Autoencoder | Autoencoders work similarly to word embeddings since they provide a dense feature representation of the input data. However, they are not limited to natural language data but can be applied to any type of input. Such architectures usually consist of an encoding stage where the input is compressed into a low-dimensional representation and a decoding stage in which the network tries to reconstruct the original input from the learned features. In this way, the network is forced to keep meaningful information in the latent representation while disregarding irrelevant noise (Goodfellow et al. 2016). Autoencoders are commonly applied for unsupervised feature learning and dimensionality reduction in combination with other subsequent learning systems. However, due to their capability of quantifying reconstruction errors, which are assumed to be significantly higher for anomalous samples than for regular instances, they can also be applied for detecting anomalies, such as fraudulent activities in financial markets (Paula et al. 2016). |
| Generative adversarial neural network (GAN) | *Generative adversarial neural networks* belong to the family of generative models that aim at learning a probability distribution over a set of training data so that the network can randomly generate new data samples with some variation. For this purpose, GANs consist of two competing sub-networks. The first network is a generator network that captures the distribution of the input and generates new examples. The second network is a discriminator network trying to distinguish real examples from artificially generated ones. Both networks are trained together in a non-cooperative zero-sum game where one network's gain is another one's loss until the discriminator can no longer distinguish between both types of samples. On this basis, GANs are likely to revolutionize domains in which continuously new content or novel product configurations are created (e.g., the composition of art and music, design of fashion), or where content is converted from one representation to another (e.g., text to image for product descriptions) (Pan et al. 2019). At the same time, however, such approaches also pose severe threats with societal implications when abusing them for malicious purposes. In particular, the generation of "deepfake" content in the form of abusive speeches and misleading news to manipulate public opinions or distort financial markets is concerning (Westerlund 2019). |

To identify a suitable prediction model for a specific task, it is reasonable to compare alternative models of varying complexities, that is, considering competing model classes as well as alternative variants of the same model class. As introduced above, a model's complexity can be characterized by several properties such as the type of learning mechanisms (e.g., shallow ML vs. DL), the number and type of manually generated or self-extracted features, and the number of trainable parameters (e.g., network weights in ANNs). Simpler models usually do not tend to be flexible enough to capture (non-linear) regularities and patterns that are relevant for the learning task. Overly complex models, on the other hand, entail a higher risk of overfitting. Furthermore, their reasoning is more difficult to interpret (cf. next section), and they are likely to be computationally more expensive. Computational costs are expressed by memory requirements and the inference time to execute a model on new data. These criteria are particularly important when assessing deep neural networks, where several million model parameters may be processed and stored, which places special demands on hardware resources. Consequently, it is crucial for business settings with limited resources (such as environments that heavily rely on mobile devices) to not only select a model at the sweet spot between underfitting and overfitting. They should also to evaluate a model's complexity concerning further trade-off relationships, such as accuracy vs. memory usage and speed (Heinrich et al. 2019).

## Challenges for intelligent systems based on machine learning and deep learning

Electronic markets are at the dawn of a technology-induced shift towards data-driven insights provided by intelligent systems (Selz 2020). Already today, shallow ML and DL are used to build analytical models for them, and further diffusion is foreseeable. For any real-world application, intelligent systems do not only face the task of model building, system specification, and implementation. They are prone to several issues rooted in how ML and DL operate, which constitute challenges relevant to the Information Systems community. They do require not only technical knowledge but also involve human and business aspects that go beyond the system's confinements to consider the circumstances and the ecosystem of application.

## Managing the triangle of architecture, hyperparameters, and training data

When building shallow ML and DL models for intelligent systems, there are nearly endless options for algorithms or architectures, hyperparameters, and training data (Duin 1994; Heinrich et al. 2021). At the same time, there is a lack of established guidelines on how a model should be built for a specific problem to ensure not only performance and cost-efficiency but also its robustness and privacy. Moreover, as outlined above, there are often several trade-off relations to be considered in business environments with limited resources, such as prediction quality vs. computational costs. Therefore, the task of analytical model building is the most crucial since it also determines the business success of an intelligent system. For example, a model that can perform at 99.9% accuracy but takes too long to put out a classification decision is rendered useless and is equal to a 0%-accuracy model in the context of time-critical applications such as proactive monitoring or quality assurance in smart factories. Further, different implementations can only be accurately compared when varying only one of the three edges of the triangle at a time and reporting the same metrics. Ultimately, one should consider the necessary skills, available tool support, and the required implementation effort to develop and modify a particular DL architecture (Wanner et al. 2020).

Thus, applications with excellent accuracy achieved in a laboratory setting or on a different dataset may not translate into business success when applied in a real-world environment in electronic markets as other factors may outweigh the ML model's theoretical achievements. This implies that researchers should be aware of the situational characteristics of a models' real-world application to develop an efficacious intelligent system. It is needless to say that researchers cannot know all factors a priori, but they should familiarize themselves with the fact that there are several architectural options with different baseline variants, which suit different scenarios, each with their characteristic properties. Furthermore, multiple metrics such as accuracy and $F1$ score should be reviewed on consistent benchmarking data across models before making a choice for a model.

## Awareness of bias and drift in data

In terms of automated analytical model building, one needs to be aware of (cognitive) biases that are introduced into any shallow ML or DL model by using human-generated data. These biases will be heavily adopted by the model (Fuchs 2018; Howard et al. 2017). That is, the models will exhibit the same (human-)induced tendencies that are present in the data or even amplify them. A cognitive bias is an illogical inference or belief that individuals adopt due to flawed reporting of facts or due to flawed decision heuristics (Haselton et al. 2015). While data-introduced bias is not a particularly new concept, it is amplified in the context of ML and DL if training data has not been properly selected or pre-processed, has class imbalances, or when inferences are not reviewed responsibly. Striking examples include Amazon's AI recruiting software that showed discrimination against women or Google's Vision AI that produced starkly different image labels based on skin color.

Further, the validity of recommendations based on data is prone to concept drift, which describes a scenario, where "the relation between the input data and the target variable changes over time" (Gama et al. 2014). That is, ML models for intelligent systems may not produce satisfactory results, when historical data does not describe the present situation adequately anymore, for example due to new competitors entering a market, new production capabilities becoming available, or unprecedented governmental restrictions. Drift does not have to be sudden but can be incremental, gradual, or reoccurring (Gama et al. 2014) and thus hard to detect. While techniques for automated learning exist that involve using trusted data windows and concept descriptions (Widmer and Kubat 1996), automated strategies for discovering and solving business-related problems are a challenge (Pentland et al. 2020).

For applications in electronic markets, considering bias is of high importance as most data points will have human points of contact. These can be as obvious as social media posts or as disguised as omitted variables. Further, poisoning attacks during model retraining can be used to purposefully insert deviating patterns. This entails that training data needs to be carefully reviewed for such human prejudgments. Applications based on this data should be understood as inherently biased rather than as impartial AI. This implies that researchers need to review their datasets and make public any biases they are aware of. Again, it is unrealistic to assume that all bias effects can be explicated in large datasets with high-dimensional data. Nevertheless, to better understand and trust an ML model, it is important to detect and highlight those effects that have or may have an impact on predictions. Lastly, as constant drift can be assumed in any real-world electronic market, a trained model is never finished. Companies must put strategies in place to identify, track, and counter concept drift that impacts the quality of their intelligent system's decisions. Currently, manual checks and periodic model retraining prevail.

## Unpredictability of predictions and the need for explainability

The complexity of DL models and some shallow ML models such as random forest and SVMs, often referred to as of black-box nature, makes it nearly impossible to predict how they will perform in a specific context (Adadi and Berrada 2018). This also entails that users may not be able to review and understand the recommendations of intelligent systems based on these models. Moreover, this makes it very difficult to prepare for adversarial attacks, which trick and break DL models (Heinrich et al. 2020). They can be a threat to high-stake applications, for example, in terms of perturbations of street signs for autonomous driving (Eykholt et al. 2018). Thus, it may become necessary to explain the decision of a black-box model also to ease organizational adoption. Not only do humans prefer simple explanations to trust and adopt a model, but the requirement of explainability may even be enforced by law (Miller 2019).

The field of explainable AI (XAI) deals with the augmentation of existing DL models to produce explanations for output predictions. For image data, this involves highlighting areas of the input image that are responsible for generating a specific output decision (Adadi and Berrada 2018). Concerning time series data, methods have been developed to highlight the particular important time steps influencing a forecast (Assaf and Schumann 2019). A similar approach can be used for highlighting words in a text that lead to specific classification outputs.

Thus, applications in electronic markets with different criticality and human interaction requirements should be designed or augmented distinctively to address the respective concerns. Researchers must review the applications in particular of DL models for their criticality and accountability. Possibly, they must choose an explainable white-box model over a more accurate black-box model (Rudin 2019) or consider XAI augmentations to make the model's predictions more accessible to its users (Adadi and Berrada 2018).

## Resource limitations and transfer learning

Lastly, building and training comprehensive analytical models with shallow ML or DL is costly and requires large datasets to avoid a cold start. Fortunately, models do not always have to be trained from scratch. The concept of transfer learning allows models that are trained on general datasets (e.g., large-scale image datasets) to be specialized for specific tasks by using a considerably smaller dataset that is problem-specific (Pouyanfar et al. 2019). However, using pre-trained models from foreign sources can pose a risk as the models can be subject to biases and adversarial attacks, as introduced above. For example, pre-trained models may not properly reflect certain environmental constraints or contain backdoors by inserting classification triggers, for example, to misclassify medical images (Wang et al. 2020). Governmental interventions to redirect or suppress predictions are conceivable as well. Hence, in high-stake situations, the reuse of publicly available analytical models may not be an option. Nevertheless, transfer learning offers a feasible option for small and medium-sized enterprises to deploy intelligent systems or enables large companies to repurpose their own general analytical models for specific applications.

In the context of transfer learning, new markets and ecosystems of AI as a service (AIaaS) are already emerging. Such marketplaces, for example by Microsoft or Amazon Web Services, offer cloud AI applications, AI platforms, and AI infrastructure. In addition to cloud-based benefits for deployments, they also enable transfer learning from already established models to other applications. That is, they allow customers with limited AI development resources to purchase pre-trained models and integrate them into their own business environments (e.g., NLP models for chatbot applications). New types of vendors can participate in such markets, for example, by offering transfer learning results for highly domain-specific tasks, such as predictive maintenance for complex machines. As outlined above, consumers of servitized DL models in particular need to be aware of the risks their black-box nature poses and establish similarly strict protocols as with human operators for similar decisions. As the market of AIaaS is only emerging, guidelines for responsible transfer learning have yet to be established (e.g., Amorós et al. 2020).

## Conclusion

With this fundamentals article, we provide a broad introduction to ML and DL. Often subsumed as AI technology, both fuel the analytical models underlying contemporary and future intelligent systems. We have conceptualized ML, shallow ML, and DL as well as their algorithms and architectures. Further, we have described the general process of automated analytical model building with its four aspects of data input, feature extraction, model building, and model assessment. Lastly, we contribute to the ongoing diffusion into electronics markets by discussing four fundamental challenges for intelligent systems based on ML and DL in real-world ecosystems.

Here, in particular, AIaaS constitutes a new and unexplored electronic market and will heavily influence other established service platforms. They will, for example, augment the smartness of so-called smart services by providing new ways to learn from customer data and provide advice or instructions to them without being explicitly programmed to do so. We estimate that much of the upcoming research on electronic markets will be against the backdrop of AIaaS and their ecosystems and devise new applications, roles, and business models for intelligent systems based on DL. Related future research will need to address and factor in the challenges we presented by providing structured methodological guidance to build analytical models, assess data collections and model performance, and make predictions safe and accessible to the user.

**Funding** This research and development project is funded by the Bayerische Staatsministerium für Wirtschaft, Landesentwicklung und Energie (StMWi) within the framework concept "Informations- und Kommunikationstechnik" (grant no. DIK0143/02) and managed by the project management agency VDI+VDE Innovation + Technik GmbH..

Machine learning and deep learning

**Publisher's note** Springer Nature remains neutral with regard to jurisdictional claims in published maps and institutional affiliations.